%% file: main.tex
\documentclass[runningheads]{llncs}
\usepackage{graphicx}
\usepackage{hyperref}
\usepackage{caption}
\usepackage{subcaption}
\hypersetup{
    colorlinks=true,
    linkcolor=blue,
    filecolor=magenta,      
    urlcolor=cyan,
    }
\begin{document}
\title{Image Contrast Enhancement using Fuzzy Technique with Parameter Determination using Metaheuristics}
%
%\titlerunning{Abbreviated paper title}
% If the paper title is too long for the running head, you can set
% an abbreviated paper title here
%
\author{Mohimenul Kabir\inst{1} \and
Jaiaid Mobin\inst{2} \and
 Ahmad Hassanat\inst{3} \and 
M. Sohel Rahman\inst{2}\thanks{Corresponding Author: \url{msrahman@cse.buet.ac.bd}}}
\authorrunning{Kabir et al.}
\titlerunning{Image Contrast Enhancement}
% First names are abbreviated in the running head.
% If there are more than two authors, 'et al.' is used.
%
\institute{
National University of Singapore
\and
Bangladesh University of Engineering \& Technology \and
Mutah University}
\maketitle              % typeset the header of the contribution
\begin{abstract}
In this work, we have presented a way to increase the contrast of an image. Our target is to find a transformation that will be image specific. We have used a fuzzy system as our transformation function. To tune the system according to an image, we have used Genetic Algorithm and Hill Climbing in multiple ways to evolve the fuzzy system and conducted several experiments. Different variants of the method are tested on several images and two variants that are superior to others in terms of fitness are selected. We have also conducted a survey to assess the visual improvement of the enhancements made by the two variants. The survey indicates that one of the methods can enhance the contrast of the images visually.

\end{abstract}
\input{section/introduction.tex}
\input{section/literature_review.tex}
\input{section/preliminaries.tex}
\input{section/problem_definition.tex}

\input{section/methodology.tex}

\input{section/experiments.tex}
\input{section/conclusion.tex}

\end{document}

%% file: section/introduction.tex
\section{Introduction}
Image enhancement is the procedure of improving an image's quality and information content. Image enhancement aims to increase visual differences among its features and make it more suitable for applications (e.g. increasing the brightness of dark images for viewing). Some common image enhancement techniques are sharpening, smoothing, increasing contrast, noise reduction, etc .~\cite {draa2014artificial}. Contrast enhancement is a process that is applied to images or videos to increase their dynamic range \cite{hashemi2010image}.

Image enhancement has been practiced as an applicable problem in meta-heuristics for a long time~\cite{cuckoo_search_and_PSO,ACO_GA_SA,munteanu2004gray}. Moreover, the blending of fuzzy systems with meta-heuristic algorithms has recently received attention in the
Computational Intelligence community \cite{fuzzy_plus_meta}. Many works treated the image enhancement problem as an optimization problem and concentrated on altering the image quality fitness function \cite{cuckoo_search_and_PSO,draa2014artificial,oloyede2019new}, combining several meta-heuristic algorithms \cite{cuckoo_search_and_PSO,ACO_GA_SA}, optimizing parameters \cite{samanta2018log} and escaping local optima \cite{dos2009differential}. One thing that has yet to gain more attention in image enhancement problems is the adaption of a fuzzy system. Sandeep and Samrudh \cite{s2018image} have shown that variation in the input membership function in a fuzzy system has a positive impact on image enhancement performance.

This paper addresses the image contrast enhancement problem by using stochastic optimization on the fuzzy logic system. The paper's main contribution is to design a  meta-heuristic technique by optimizing the fuzzy logic system. The logical component of the fuzzy logic system is a set of membership functions that are used to describe an intensity transformation function. We implement genetic operators which tweak the fuzzy logic system to enhance the original image to an optimal enhanced image. The fuzzy system of our paper is based on simple contrast enhancement rules described in \cite{gonzalez2008digital}, where the authors have used some image-independent rules and fuzzy sets. 

Our main idea of the fuzzy image enhancement technique is as follows: we start with a basic fuzzy rule set and a set of input membership functions. By applying a metaheuristics framework, we try to evolve the fuzzy sets. Finally, we apply the transformation function described by fuzzy sets on the value channel of HSV color space to get the final image. Thus we have converted the problem into an optimization problem. In other words, rather than generating an image, we try to generate a suitable mapping between the input and output color values. However, this is inherently challenging since, even if we consider only 8-bit color, i.e., $256$ color values, the number of possible mappings becomes huge. Thus it becomes infeasible to search through the solution space exhaustively, and there is no definite knowledge about how to improve/generate a solution. This motivates us to leverage a metaheuristics framework \cite{Luke2009Metaheuristics}.

%% file: section/literature_review.tex
\section{Literature Review}
Some contrast manipulation techniques are gamma transformation~\cite{guan2009image}, histogram equalization (HE)~\cite{singh2015histogram}, etc. HE is very useful in contrast enhancement \cite{stark2000adaptive}. But HE, while increasing the contrast, fails to keep image brightness the same \cite{shanmugavadivu2014particle}. BiHistogram Equalization~\cite{ooi2009bi} solves this issue. Another problem of HE is the information loss of image \cite{zhu2012adaptive}. Also, gamma transform, log transform~\cite{singh2014various} can be used with lower computational complexity. In the work of \cite{gamma}, we can find applications of gamma transformation for contrast enhancement which applies different gamma corrections on multiple parts of pixel sets automatically. Unfortunately, these techniques don't work well in a complex illumination setting \cite{oloyede2019new}. So, they can not be applied without tweaking some parameters. 

Fuzzy logic and metaheuristics techniques have been applied previously  in image enhancement problems. One basic method is to apply a fuzzy logic-based system from \cite{gonzalez2008digital} that uses three rules only and trapezoidal and triangular input fuzzy sets. Joshi and Kumar~\cite{s2018image} have proposed a similar method, albeit with a more complex rule set ($7$ rules)  and Gaussian fuzzy sets.

To evaluate the fitness of an enhanced image, Munteanu and Rosa~\cite{munteanu2004gray} have proposed a novel objective function and applied an evolutionary algorithm to search for optimal parameters in a continuous transform function. The same objective function has also been applied in artificial bee colony optimization \cite{draa2014artificial}, cuckoo search algorithm \cite{bhandari2019cuckoo}, and also in the firefly algorithm for UAV captured image enhancement \cite{samanta2018log}. 

\iffalse
From this brief review, we can see that both the metaheuristics framework and the fuzzy logic system is a used approach in image enhancement. The limitation of current fuzzy logic-based literature is that the fuzzy logic systems used are constant logic systems. Using the metaheuristics framework, deriving a better fuzzy logic system for image enhancement can be automated.
\fi

%% file: section/preliminaries.tex
\section{Preliminaries}
\subsection*{Fuzzy Image Processing}
\begin{figure}
    \centering
    \includegraphics[width=0.8\textwidth]{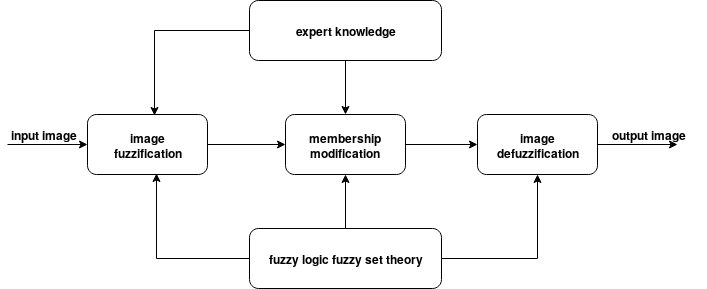}
    \caption{Fuzzy Image Processing. The figure is taken from \cite{unknown}}
    \label{fig:fii}
\end{figure}

Fuzzy image processing is a collection of all approaches that understand, represent, and process the images, their segments, and features as fuzzy  sets (see Figure \ref{fig:fii}).
The coding of  image data (fuzzification) and decoding of the results  (defuzzification)  are  steps that make it possible  to  process  images  with  fuzzy techniques. The main power of fuzzy image processing is in the middle step (membership modification) in Figure \ref{fig:fii} \cite{Fuzzy}.
\subsection*{Fuzzy Sets for Intensity Transformation}
Contrast enhancement is one of the principal
applications of intensity transformations. We can state the process of enhancing the contrast of a gray-scale image using the following rules \cite{gonzalez2008digital}:
\begin{itemize}
    \item IF a pixel is \textbf{dark}, THEN make it \textbf{darker}
    \item IF a pixel is \textbf{gray}, THEN make it \textbf{gray}
    \item IF a pixel is \textbf{bright}, THEN make it \textbf{brighter}
\end{itemize}

%% file: section/problem_definition.tex
\section{Problem Statement}
Our goal is to manipulate the image contrast to enhance the sharpness of the image. Thus, image features will be more differentiable visually. While this is a subjective matter, the effort has been made to quantify it using a fitness function in the literature \cite{bhandari2019cuckoo,draa2014artificial,dos2009differential,braik2007particle,munteanu2004gray}. We leverage one such fitness function in our work. This fitness function is used to measure the quality of an image. A transformation function is used to enhance the image, which is further optimized using a metaheuristics approach. So, the input of the problem is an image, and the output is another image which is (expectedly) an enhanced version of the former.

We give a more formal definition of the problem below in the context of a gray-scale image (for simplicity). Suppose a gray-scale image, $I = f(x,y)$, where $x$ and $y$ denote the pixels' positions of the image. Image $I$ is of $M \times N$ size. So,  $0\leq x < M$ and $0\leq y < N$. Now assume that there is a fuzzy logic-based transformation function $T(I)$ that transforms the gray value of each pixel of the image and outputs another image $I_e=g(x,y)$ and a quality function, $Fitness(I)$

\begin{equation} \label{eqn:transformation}
I_e = T(I) = T(f(x,y)) 
\end{equation}

\begin{equation} \label{eqn:fitness}
F = Fitness(I) = log(log(E(I_s)))*\frac{ne(I_s)}{M*N}*H(I_s)
\end{equation}
So, our problem is to find a transformation function (Equation \ref{eqn:transformation}) by optimizing (here maximization) Equation \ref{eqn:fitness}. For explanation of the terms in Equation \ref{eqn:fitness}, see Subsection \ref{subsection:fitness_assessment}.

%% file: section/methodology.tex
\section{Methodology}
\label{section:method}
In this section, we present our approaches. We have explored five different metaheuristic approaches. In particular, we have used Hill Climbing (three variants) and Genetic Algorithm (two variants thereof). For the Hill Climbing approach, the variants differ in the mutation functions and input membership functions, as highlighted below. We used one type of input membership function for the Genetic Algorithm, but the difference is in new generation selection strategies (see Section \ref{subsection:method_ga_join}).
\begin{itemize}
    \item Hill Climbing
        \begin{enumerate}
            \item neighborhood generation using simple mutation
            \item neighborhood generation with trapezoidal and triangular input membership set splitting mutation
            \item mixed neighborhood generation with Gaussian input membership set splitting mutation
        \end{enumerate}
    \item Genetic Algorithm
    \begin{enumerate}
            \item simple neighborhood generation with trapezoidal and triangular input membership set
            \item simple neighborhood generation with gaussian and sigmoid input membership set
        \end{enumerate}
\end{itemize}
All these variants have some common parts, namely, initialization, representation, and fitness assessment of an optimization session, described below.
\begin{figure}
    \centering
    \includegraphics[scale=0.8]{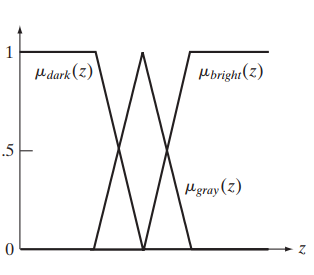}
    \caption{Input membership functions for fuzzy, rule-based contrast enhancement  \cite{gonzalez2008digital}}
    \label{fig:fuzzy_function}
\end{figure}

Finally, we have presented our methodology of experiments and values of parameters used in algorithms and experiments.

%\subsection*{Population Initialization}
\subsection{Population Representation}
\label{subsection:method_pop}
Every meta-heuristics technique starts with some (usually random) population, and the structure of the population is problem specific. In our case, each individual (in the population) will be one of the input membership functions, as shown in Figure \ref{fig:fuzzy_function}. So, our population representation holds information on a set of input membership functions. There are at least three functions in each representation. Each membership function is represented by a tuple of three values regardless of the function types (i.e., trapezoidal, triangular, Gaussian, and Sigmoidal). The first two values of the tuple determine the shape of the membership function, and the third one is used in defuzzification (as will be clear shortly).

\subsection{Fitness Assessment}
\label{subsection:fitness_assessment}
We will formulate the enhancement's quality/fitness $F$ using the quality function mentioned in Equation \ref{eqn:fitness}. There are some terms in Equation \ref{eqn:fitness} which we will explain now, 

\begin{description}
    \item[$I_s$] $=$ image after Applying Sobel filter on $I$
    \item[$E(I)$] $=$ sum of intensity of image $I$
    \item[$ne(I)$] $=$ number of edge pixel in $I$
    \item[$H(I)$] $=$ entropy of image $I$
    \item[$M$] $=$ width of $I$
    \item[$N$] $=$ height of $I$
\end{description}

\subsection*{HSV Conversion}
\label{subsection:method_hsv}
If the input image is a color image, we need to convert it to the HSV format. In the case of a gray-scale image, there is no need for such a conversion. 

\subsection*{Defuzzification}
\label{subsection:method_defuzz}
Defuzzification refers to converting the fuzzy value to a single crisp value. For any given input pixel $z_0$, the output crisp value $v_0$ will be as follows.
\begin{equation} \label{eqn:defuzzy}
v_0 = \frac{\sum\limits_{i=1}^n\mu_{i}(z_0) * v}{\sum\limits_{i=1}^n\mu_{i}(z_0)};n\geq3
\end{equation}
Here,
\begin{description}
    \item[$\mu_{i}(z_0) =$] fuzzy level of the image pixel with value $z_0$
    \item[$v =$] constant/mutable value (depending on the algorithm variant) used in the defuzzification 
\end{description}

\subsection{Hill Climbing}
\label{subsection:method_hc}
Hill Climbing stochastically generates candidate solutions and stores the best-found solution. The best solution is evaluated using the fitness function shown in Equation \ref{eqn:fitness}. In each generation, we generate a certain number of neighbor solutions (set to $10$) from a fixed individual. We have tried three variants of Hill Climbing differing from each other in neighborhood generation and input membership function type.

\subsubsection{Simple Neighborhood Generation}
\label{subsubsec:simple_neighbour_generation}
In this variant, we have initiated a solution with two trapezoidal ($\mu\_dark, \mu\_bright$) and one triangular ($\mu\_gray$) functions like Figure \ref{fig:fuzzy_function}. When generating a new solution, only the shape of the functions (i.e., the width of the triangle and trapezoid's oblique line's slope) are tweaked. We have used three hyperparameters, namely, $\mathsf{ChangeProb}$, $\mathsf{MutateMu}$, and $\mathsf{MutateSigma}$. Firstly, $\mathsf{ChangeProb}$ represents the threshold of the random number chosen to decide whether the shape change of a member function will be done. On the other hand, $\mathsf{MutateMu}$ and $\mathsf{MutateSigma}$ represent, respectively, the mean and variance of Gaussian distribution from which the random number is chosen when generating a new solution. 
\subsubsection*{Neighborhood Generation with trapezoidal and triangular input membership set splitting mutation}
\label{subsubsection:method_hc_traptri_split}
This variant considers an additional tweak besides the previous one. A membership function may split into two membership functions (e.g., one triangle in Figure \ref{fig:fuzzy_function} splits into two). In addition to the previously mentioned parameters, here we have used another one, which is $\mathsf{MembershipSplitProb}$, to control the probability of choosing function shape change or input member function splitting.
\subsubsection*{Neighborhood Generation with Gaussian input membership set splitting mutation}
\label{subsubsection:method_hc_gauss_split}
In this case, we have used only the Gaussian input membership function in our solution. All the hyperparameters mentioned in the previous two cases are used.

\subsection{Genetic Algorithm}
\label{subsection:method_ga}
Unlike Hill Climbing, the Genetic Algorithm (GA) starts with a number (say, $\mathsf{PopSize}$) of individuals. We use the same evaluation function shown in Equation \ref{eqn:fitness} for fitness evaluation in our GA approach. To introduce variations in the population, GA breeds a new population of children, selects individuals from the old population, and tweaks them to breed new individuals. To keep the footprints of both populations, it joins the parent and children populations to form a new generation of population. In our implementation, we have fixed $\mathsf{PopSize}$ equal to $30$.

\subsection{Tweaking Operations}
\label{subsection:method_ga_tweak}
\subsubsection{Crossover}
The crossover operation mixes multiple individuals (typically $2$) and matches to form the children. There are $3$ classical ways of doing a crossover. In our implementation, we have used uniform crossover. In the context of our representation, our crossover operation marches down all the membership functions and to combine them swaps individual functions if a coin toss comes up head with probability $p$. To use crossover on the Genetic Algorithm, both individuals should be of the same size. 

\subsubsection{Mutation}
\label{subsection:method_ga_mut}
Our mutation operation scans each membership function and randomly tweaks the function shape with a certain probability. To exploit both exploitation and exploration, our mutation operation uses a Gaussian distribution with a certain mean and variance. Unlike Hill Climbing, mutation operation here does not increase/decrease the number of membership functions.

\subsection{Joining}
\label{subsection:method_ga_join}
The Genetic Algorithm differs in how parent and child populations are joined. In our Genetic Algorithm procedure, we have experimented with both (P, P) and (P+P) evolution strategies, where $P$ is the $\mathsf{PopSize}$.

\subsection{Experiment Process}
\label{subsection:experiment_process}
In the experimental analysis, first, we choose the best variants among all variants of Hill Climbing and the Genetic Algorithm. Then we measure the performance of our image enhancement method with respect to common metrics of image enhancements, and finally, we compare our results with one of the simplest methods of image enhancement --- histogram equalization.
%The objective of our experiment is first select the best two variants from the five variants we have experimented on, then we have measured different metrics of the output images to compare them with a simple contrast enhancement approach called histogram equalization.

\subsubsection{Variant Selection}
The following steps are used to choose the best variants:

\begin{enumerate}
    \item Run one variant on each image. The variant is applied on one image a total of $\mathsf{NumofTest}$ times. Each time (one stochastic optimization) is run for $ \mathsf{PerRunTime}$ seconds. This time is kept the same for all variants. Also, mutation and population initialization-related parameters are kept the same for all variants. 
    %(except $SPLIT\_PROB$ because variants are also differentiated based on this).  
    
    \item Measure the average rate of fitness improvement over the number of generations achieved in given $\mathsf{PerRunTime}$. Then the average is computed for $\mathsf{NumofTest}$ times.
    
    \item We have compared the metric achieved in step $2$ for all five variants and then selected the best two variants. The higher the value of this metric demonstrates better performance.
\end{enumerate}

The value of experiment controlling parameters is written in Table \ref{tab:exp_parameters}.

\begin{table}[]
    \centering
    \begin{tabular}{| c | c |}
        \hline
        \textbf{Hyperparameter name} & \textbf{Value}\\
        \hline
        $\mathsf{NumofTest}$ & $5$ \\ 
        \hline
        $\mathsf{PerRunTime}$ & $120$ \\
        \hline
        $\mathsf{ChangeProb}$ &  $0.5$\\
        \hline
        $\mathsf{MutateMu}$ & $3$ \\
        \hline
        $\mathsf{MutateSigma}$ & $2$ \\
        \hline
        $\mathsf{MembershipSplitProb}$ & $0.1$ \\
        \hline
    \end{tabular}
    \caption{The value of hyperparameters}
    \label{tab:exp_parameters}
\end{table}

%% file: section/experiments.tex
 \section{Experimental Results}
In this section, we evaluate our proposed approach and present the results of the application of our method on different images. We have conducted our experiments on several images used in  \cite{draa2014artificial} and some other images (both color, gray-scale, and text images) collected from different sources from the Internet. In particular, we have experimented with a total of $18$ images.

\subsection{Experimental Setup and Environment}
To evaluate the efficacy of our approach, we have implemented a prototype in python. We have used the python evolutionary computation framework,  deap\footnote{\url{https://deap.readthedocs.io/en/master/}}. To implement fuzzy logic, we have used python fuzzy logic toolbox, skfuzzy\footnote{\url{https://pythonhosted.org/scikit-fuzzy/}}.

All variants of our algorithm are run on a Intel(R) Core(TM) i$3$ CPU M$370@2.40$GHz processor with $6$GB RAM running ubuntu $18.04$ and python version $3.6.8$.

Our experiment code can be found in this link, \url{https://bitbucket.org/mahi045/image-enhancement/}.

\subsection{Results}
\label{subsection:exp_result_result}

\begin{figure}
\begin{subfigure}{0.5\textwidth}%
  \includegraphics[width=\linewidth]{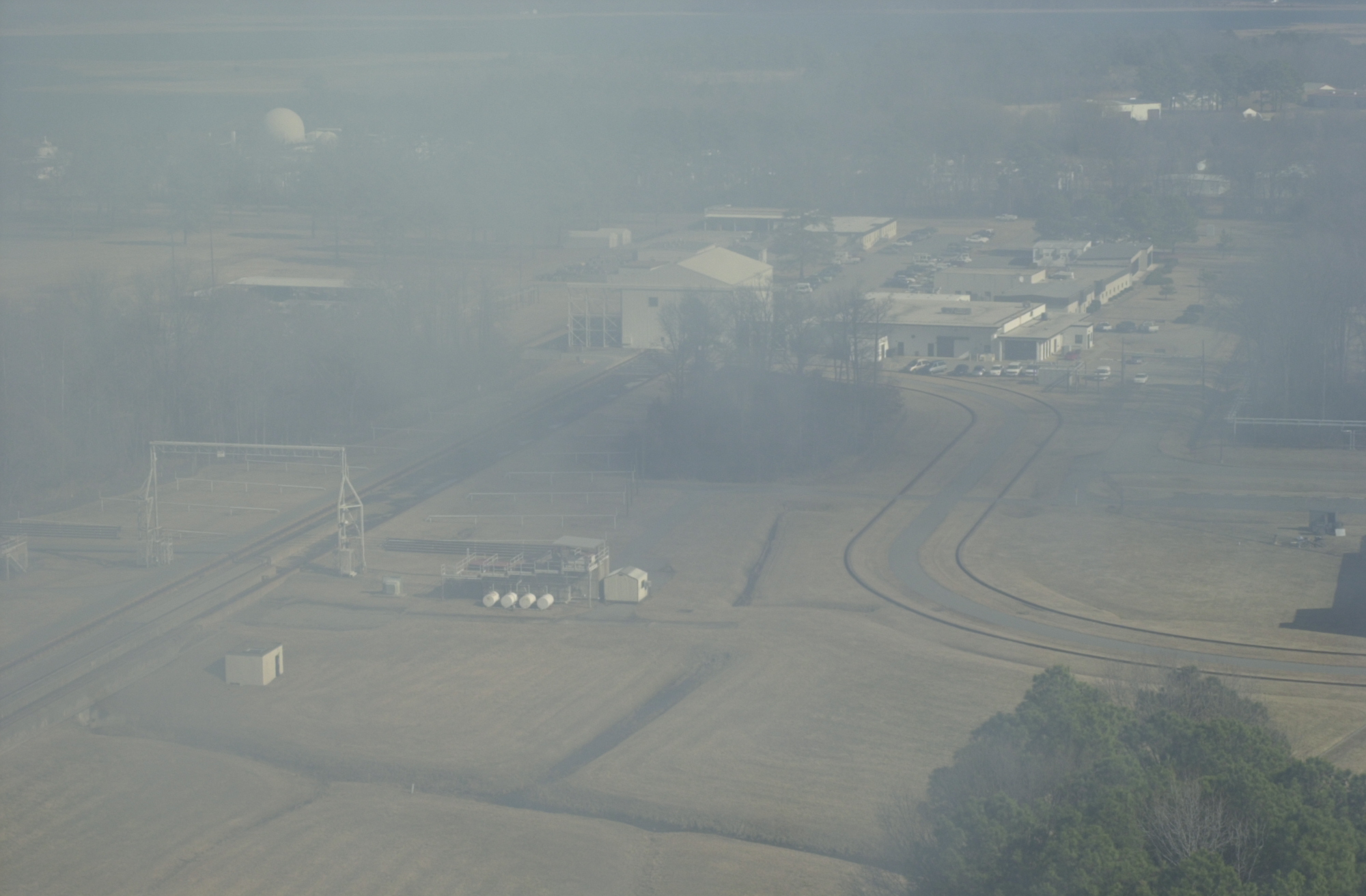}
  \caption{}
  \label{fig:04_image}
\end{subfigure}
\begin{subfigure}{0.5\textwidth}%
  \includegraphics[width=\linewidth]{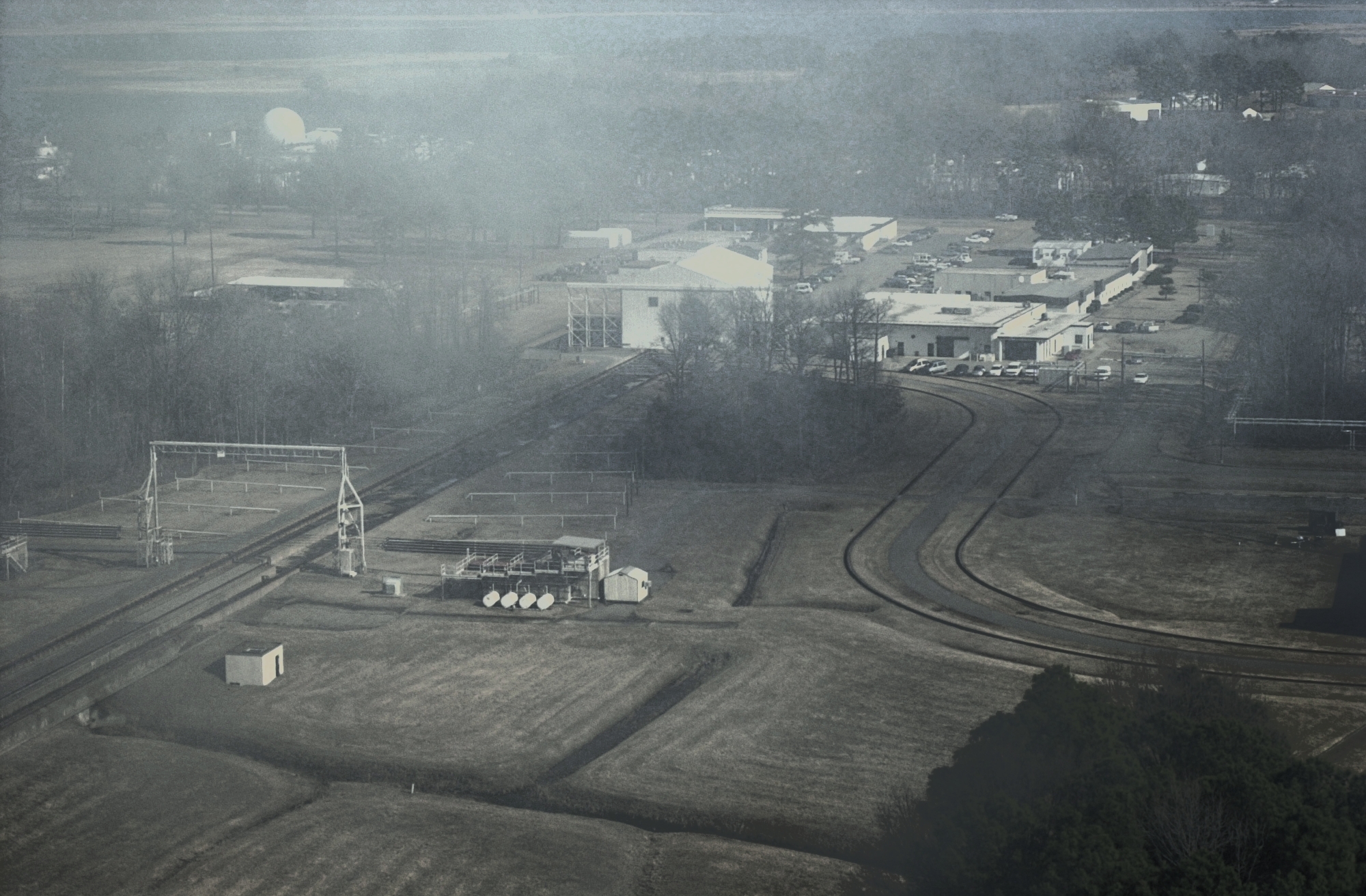}
  \caption{}
  \label{fig:04_output_image}
\end{subfigure}
  \caption{enhanced image (b) generated by simple HC and (a) is original image}
  \label{fig:04_output_simple_hc}
\end{figure}

\begin{figure}
\begin{subfigure}{0.5\textwidth}%
  \includegraphics[width=\linewidth]{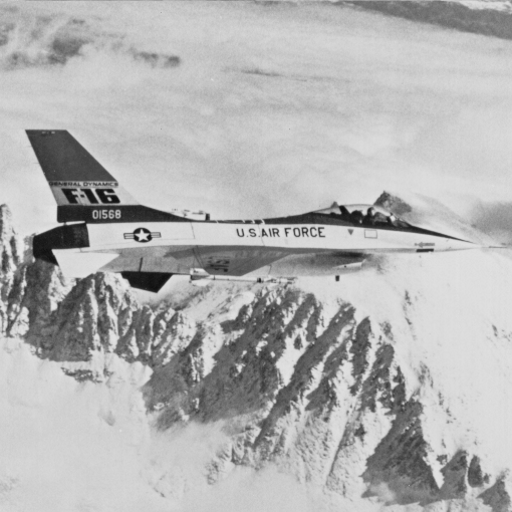}
  \caption{}
  \label{fig:04_image}
\end{subfigure}
\begin{subfigure}{0.5\textwidth}%
  \includegraphics[width=\linewidth]{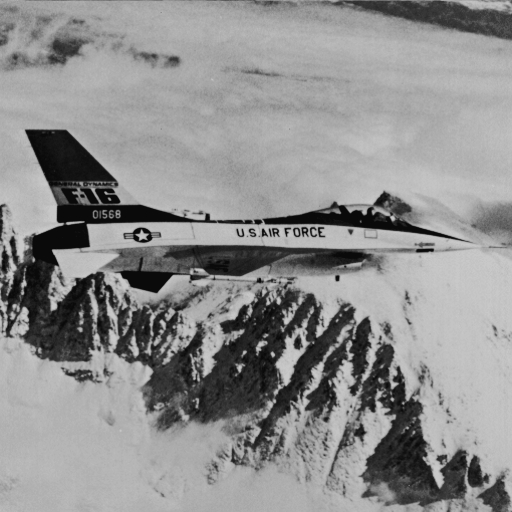}
  \caption{}
  \label{fig:04_output_image}
\end{subfigure}
  \caption{enhanced image (b) generated by simple HC and (a) is original image}
  \label{fig:jetplane_output_simple_hc}
\end{figure}

\begin{figure}
\begin{subfigure}{0.5\textwidth}
  \includegraphics[width=\linewidth]{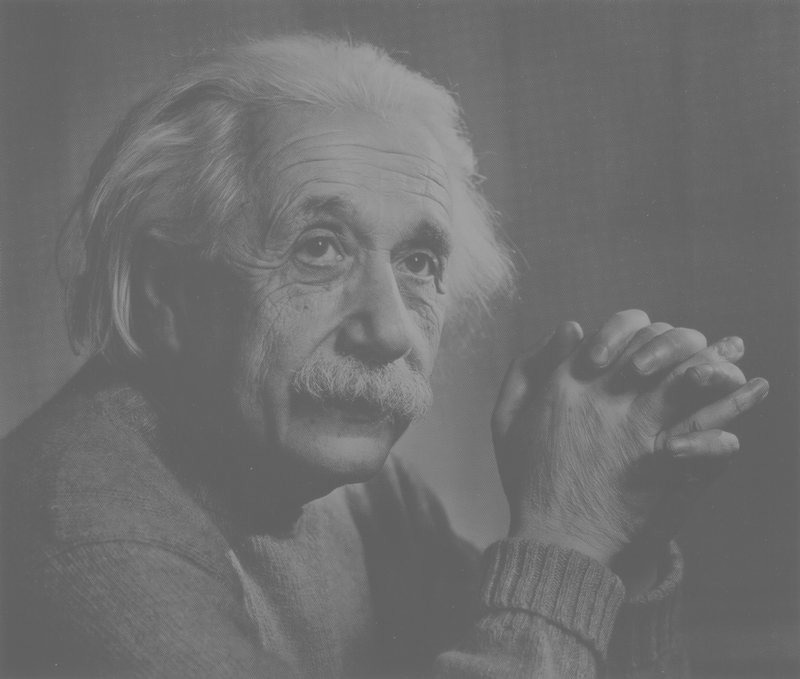}
  \caption{}
  \label{fig:01_image}
\end{subfigure}
\begin{subfigure}{0.5\textwidth}
  \includegraphics[width=\linewidth]{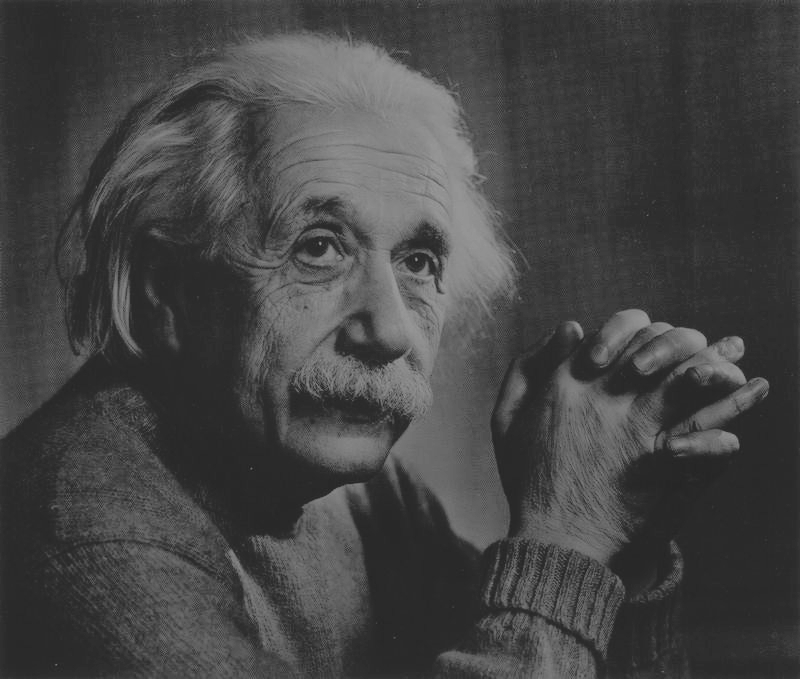}
  \caption{}
  \label{}
\end{subfigure}
\caption{enhanced image (b) generated by Genetic Algorithm (P, P) and (a) is original image}
  \label{fig:07_output_genetic_explore}
\end{figure}

\begin{figure}
\begin{subfigure}{0.5\textwidth}%
  \includegraphics[width=\linewidth]{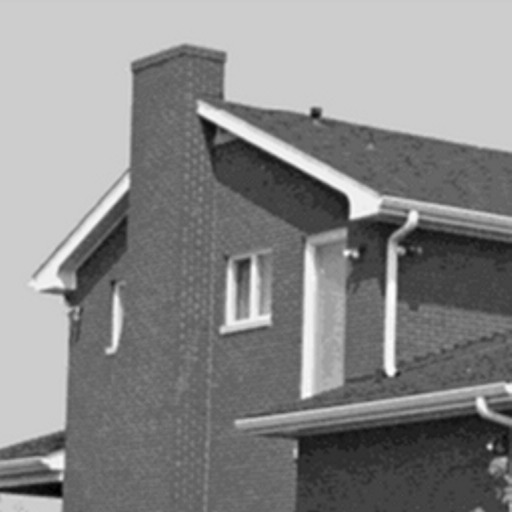}
  \caption{}
  \label{fig:03_image}
\end{subfigure}
\begin{subfigure}{0.5\textwidth}%
  \includegraphics[width=\linewidth]{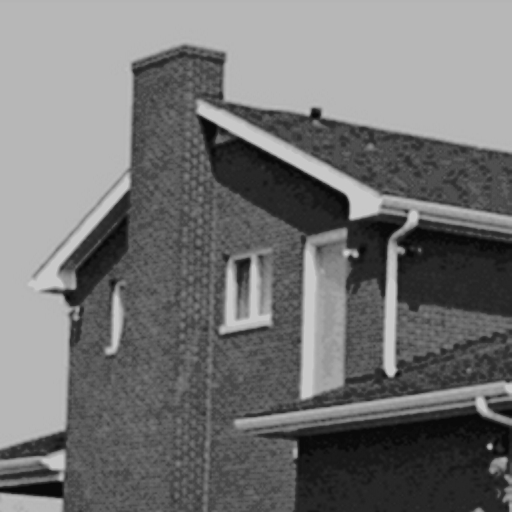}
  \caption{}
  \label{fig:03_output_image}
\end{subfigure}
\caption{enhanced image (b) generated by Genetic Algorithm (P + P) and (a) is original image}
  \label{fig:house_output_genetic_exploit}
\end{figure}

Here, we visually present outputs for a limited number of images due to our page limitation. Figures \ref{fig:04_output_simple_hc}, \ref{fig:jetplane_output_simple_hc}, \ref{fig:07_output_genetic_explore}, \ref{fig:house_output_genetic_exploit} show output images generated by our technique against the reference input images.
As an outcome of our technique, the enhanced images are more clear and natural (Figures \ref{fig:04_output_simple_hc}, \ref{fig:07_output_genetic_explore}, \ref{fig:house_output_genetic_exploit}), the edges are sharper(Figures \ref{fig:jetplane_output_simple_hc},
\ref{fig:house_output_genetic_exploit}), the histogram is expanded, and contrast between black and white portion is increased. On the contrary, some black portions (Figures \ref{fig:jetplane_output_simple_hc}, \ref{fig:07_output_genetic_explore}) are unnecessarily darkened. 

\subsection{Survey}
\label{subsection:exp_result_survey}
\subsubsection{Variant Selection to Select Survey Images}
First we have grouped the variants according to used metaheuristic algorithms, namely Hill Climbing and Genetic algorithm. Then according to the method described in Subsection \ref{subsection:experiment_process}, we have ranked the variants among two groups. Based on this the selected variants from the two groups are,
\begin{enumerate}
    \item Hill Climbing with simple neighborhood generation (Subsection \ref{subsubsec:simple_neighbour_generation})
    \item Genetic Algorithm with (P, P) strategy (Subsection \ref{subsection:method_ga_join})
\end{enumerate}

\subsubsection{Survey Preparation}
We have prepared an anonymous survey\footnote{\url{https://forms.gle/9wuLKMYBShjyAaK59}} (currently not taking any responses) to get a quantitative opinion on our enhanced images. In the survey, we have placed $36$ enhanced images (produced by our technique) side-by-side with their original image. The first $18$ images are from Hill Climbing with neighborhood generation with split mutation (see Section \ref{subsubsection:method_hc_traptri_split}), and the second $18$ images are the same but from the Genetic Algorithm with (P, P) strategy (see Section \ref{subsection:method_ga_join}). Reviewer needs to mark the enhanced image on a scale of $1-9$. As the reference, the mark of the original image was $5$; therefore, a mark of $>5$ by the reviewer would mean an enhanced image, while $<5$ will mean image degradation. A mark of $5$ would mean the processed image is the same as the original visually (a subjective judgment). The prepared survey form can be found at the following link: \url{https://forms.gle/9wuLKMYBShjyAaK59}.
%\href{https://forms.gle/9wuLKMYBShjyAaK59}{here}.

\subsubsection{Survey Response \& Observations}
We have received a total of $67$ responses. From the responses, we have filtered responses with suspicious patterns like all equal marks or very high or very low marks. This filtering caused two responses to be filtered. From the remaining $65$ responses, it seems method $2$ is better (mean score $5.35$) than method $1$ (mean score $4.62$). Moreover, the method $2$ improves the original image (score $> 5$).

\begin{figure}[!htb]
\begin{subfigure}{0.5\textwidth}%
  \includegraphics[width=\linewidth]{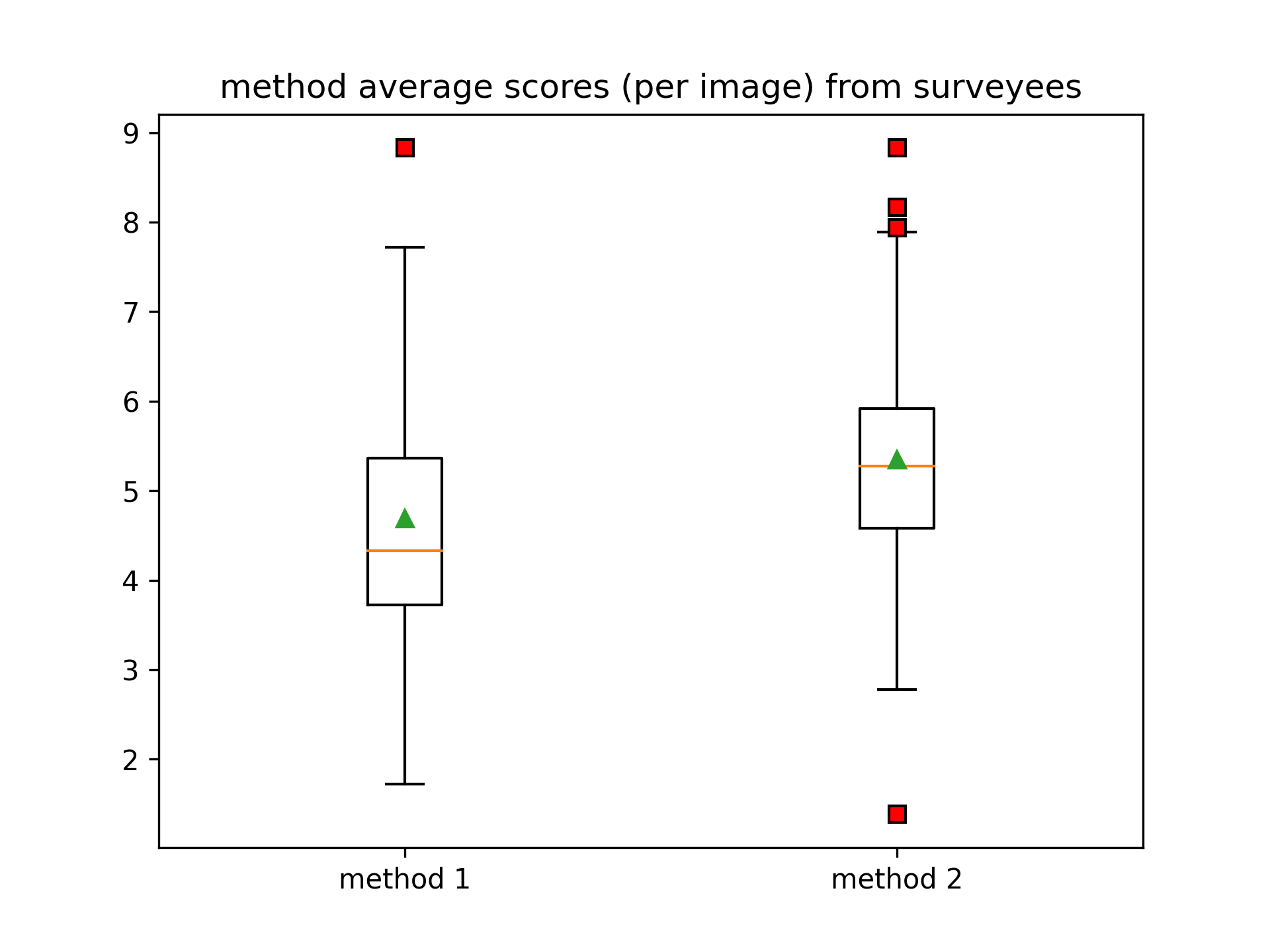}
  \caption{}
  \label{fig:survey_result_boxplot}
\end{subfigure}
\begin{subfigure}{0.5\textwidth}%
  \includegraphics[width=\linewidth]{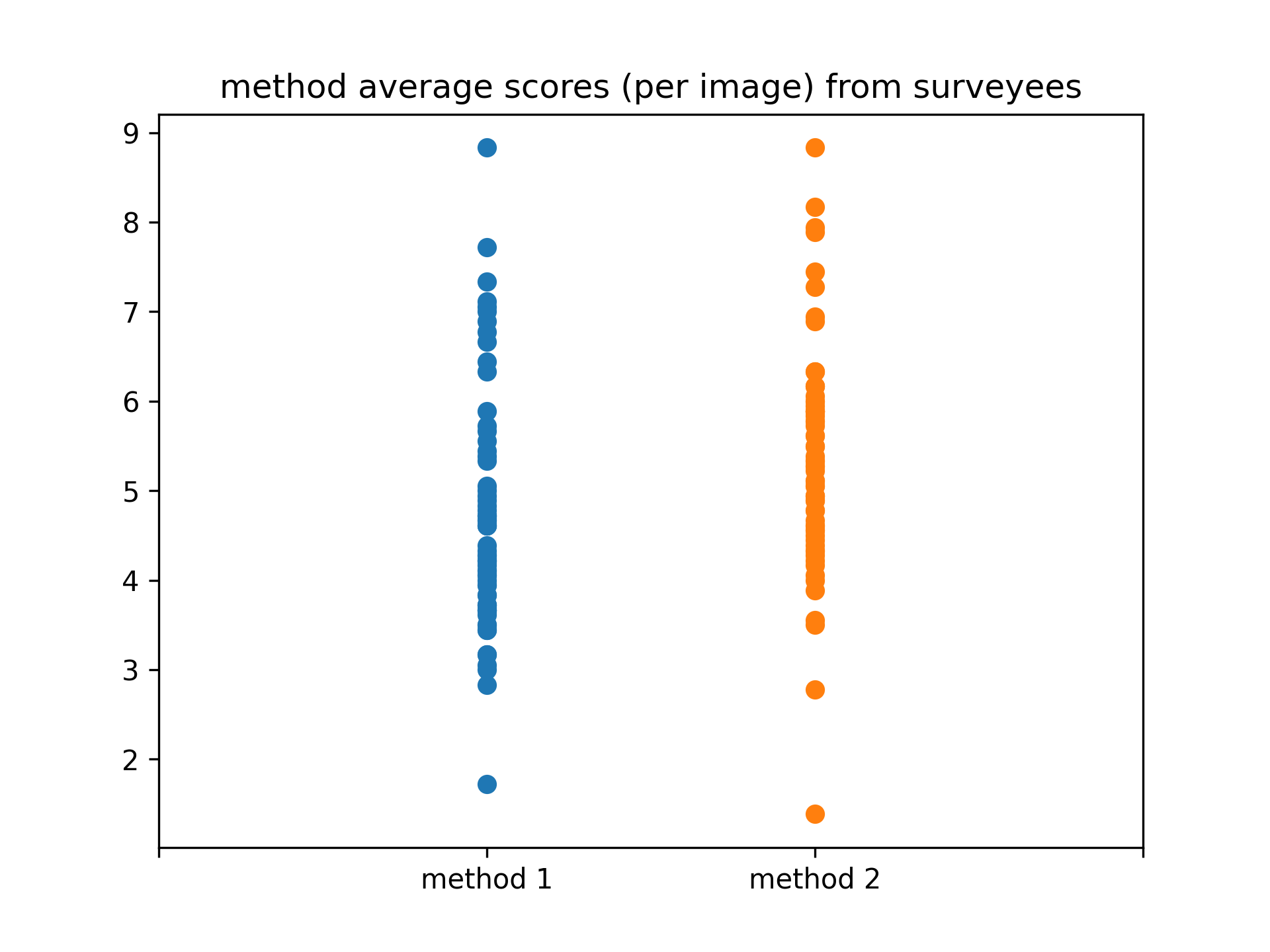}
  \caption{}
  \label{fig:survey_result_scatterplot}
\end{subfigure}
  \caption{Box plot of achieved score from survey participants of two methods. Method $1$ represents Hill Climbing with neighborhood generation with split mutation (\ref{subsubsection:method_hc_traptri_split}) and method $2$ represents Genetic Algorithm with (P, P) strategy (see Section \ref{subsection:method_ga_join}) }
  \label{fig:survey_result}
\end{figure}

From Figure \ref{fig:survey_result}, we can see the mean of (over images) quantitative responses of people (survey participants). Each data point in Figure \ref{fig:survey_result}(b) is the mean response of one participant. Figure \ref{fig:survey_result}(a) shows the mean (green triangle), and median (orange horizontal line) of the responses. The red points are outliers according to the $1.5$*IQR rule \cite{iqr}.

From Figure \ref{fig:survey_result}(a), method $2$ shows a better mean score than method $1$'s mean score (mean over participants). From Figure \ref{fig:survey_result}(b), we can see that method $1$ has a higher chance of degrading the image (dense below score $5$), although method $2$ has achieved the image with the lowest score according to one participant.


%% file: section/conclusion.tex
\section{Conclusion}
We have applied meta-heuristics to find a good image-specific fuzzy logic-based transformation function in this work. We have employed a quality function from some previous works. As image quality is subjective, to assess our method's success, we have conducted a survey to gain subjective opinions on the visual quality of enhancement and reported the result. From the assessment, we have seen that one variant among the five gives us a visual improvement on average.

For future work, more experiments can be done by associating more image processing operations (e.g., gamma transformation) can be added in the processing pipeline. One limitation of our approach (according to visual observation) is that it darkens the image. Adding other processing operations with automatic parameter tuning through meta-heuristics seems promising for further improvement.